\documentclass[conference]{IEEEtran}
\IEEEoverridecommandlockouts
\usepackage{cite}
\usepackage{amsmath,amssymb,amsfonts}
\usepackage{algorithmic}
\usepackage{graphicx}
\usepackage{textcomp}
\usepackage{xcolor}
\def\BibTeX{{\rm B\kern-.05em{\sc i\kern-.025em b}\kern-.08em
    T\kern-.1667em\lower.7ex\hbox{E}\kern-.125emX}}

\usepackage{booktabs} 
\usepackage{multirow} 

\usepackage{tabularx} % 在导言区添加
\usepackage{colortbl}      % 支持列背景色
\usepackage[table,dvipsnames]{xcolor} % 放在导言区
\begin{document}

\title{Learning Fine-Grained Geometry for Sparse-View Splatting via Cascade Depth Loss
}

\author{\IEEEauthorblockN{Wenjun Lu}
\IEEEauthorblockA{\textit{School of Computer Science} \\
\textit{The University of Sydney}\\
Sydney, Australia \\
wenjun.lu@sydney.edu.au}
\and
\IEEEauthorblockN{Haodong Chen}
\IEEEauthorblockA{\textit{School of Computer Science} \\
\textit{The University of Sydney}\\
Sydney, Australia \\
haodong.chen@sydney.edu.au}
\and
\IEEEauthorblockN{Anqi Yi}
\IEEEauthorblockA{\textit{School of Computer Science} \\
\textit{The University of Sydney}\\
Sydney, Australia \\
anqi.yi@sydney.edu.au}
\and
\IEEEauthorblockN{Guoxi Huang}
\IEEEauthorblockA{\textit{School of Computer Science} \\
\textit{University of Bristol}\\
Bristol, United Kingdom \\
guoxi.huang@bristol.ac.uk}
\and
\IEEEauthorblockN{Yuk Ying Chung}
\IEEEauthorblockA{\textit{School of Computer Science} \\
\textit{The University of Sydney}\\
Sydney, Australia \\
vera.chung@sydney.edu.au}
\and
\IEEEauthorblockN{Kun Hu}
\IEEEauthorblockA{\textit{School of Science} \\
\textit{Edith Cowan University}\\
Perth, Australia \\
k.hu@ecu.edu.au}
\and
\IEEEauthorblockN{Zhiyong Wang}
\IEEEauthorblockA{\textit{School of Computer Science} \\
\textit{The University of Sydney}\\
Sydney, Australia \\
zhiyong.wang@sydney.edu.au}
}

\maketitle

\begin{abstract}
Novel view synthesis is a fundamental task in 3D computer vision that aims to reconstruct photorealistic images from novel viewpoints given a set of posed images. However, reconstruction quality degrades sharply under sparse-view conditions due to insufficient geometric cues. Existing methods, including Neural Radiance Fields (NeRF) and more recent 3D Gaussian Splatting (3DGS), often exhibit blurred details and structural artifacts when trained from sparse observations. Recent works have identified rendered depth quality as a key factor in mitigating these artifacts, as it directly affects geometric accuracy and view consistency. However, effectively leveraging depth under sparse views remains challenging. Depth priors can be noisy or misaligned with rendered geometry, and single-scale supervision often fails to capture both global structure and fine details. To address these challenges, we introduce Hierarchical Depth-Guided Splatting (HDGS), a depth supervision framework that progressively refines geometry from coarse to fine levels. Central to HDGS is our novel Cascade Pearson Correlation Loss (CPCL), which enforces consistency between rendered and estimated depth priors across multiple spatial scales. By enforcing multi-scale depth consistency, our method improves structural fidelity in sparse-view reconstruction. Experiments on LLFF and DTU demonstrate state-of-the-art performance under sparse-view settings.
\end{abstract}

\section{Introduction}
\label{sec:introduction}

Reconstructing photorealistic images of a scene from novel viewpoints is commonly referred to as novel view synthesis (NVS). NVS supports a wide range of applications, such as virtual and augmented reality and robotic perception. These use cases often require high-quality renderings from previously unseen perspectives, making the ability to synthesize novel views both technically critical and practically valuable.

State-of-the-art NVS methods such as Neural Radiance Fields (NeRF)~\cite{mildenhall2021nerf} and the more recent 3D Gaussian Splatting (3DGS)~\cite{kerbl20233d} have shown impressive results under dense view supervision. NeRF represents the scene implicitly using multilayer perceptrons (MLPs), producing high-fidelity images but suffering from long training and inference times. In contrast, 3DGS models the scene explicitly as a set of anisotropic 3D Gaussians, allowing real-time rendering while maintaining visual quality in dense view settings.

However, in many real-world scenarios, collecting dense views is impractical, and both methods degrade sharply under sparse-view conditions. The limited observations provide insufficient geometric cues for reconstruction. NeRF’s reliance on dense photometric supervision leads to spatial ambiguity and geometric inconsistency, while 3DGS depends on reliable initialization and adequate view coverage to optimize Gaussian positions and scales, making it vulnerable to sparse inputs.

Recent approaches~\cite{zhu2024fsgs, paliwal2024coherentgs, chung2024depth, xu2024mvpgs} introduce depth priors to strengthen geometric supervision, but most perform depth alignment at a single global scale, which can miss local structural variations and lead to inaccurate scene geometry. Although some works~\cite{li2024dngaussian, xu2024mvpgs}  explore patch-level depth supervision, their objectives typically compare depth values directly rather than emphasizing
structural agreement between rendered depth and priors across resolutions. Consequently, supervision can be inconsistent across scales, limiting the model’s ability to maintain coherent
geometry from global layout down to local detail.

% instead of enforcing multi-resolution structural agreement between rendered depth and priors. 
% As a result, supervision may be inconsistent across scales, limiting geometry coherence from global layout to fine detail.

To overcome these limitations, we introduce Hierarchical Depth-Guided Splatting (HDGS), a method that enhances 3D Gaussian Splatting (3DGS) performance in sparse-view scenarios through multi-scale depth supervision. The core idea is to progressively refine scene geometry from coarse to fine levels by aligning rendered and predicted depth maps across multiple spatial resolutions. This hierarchical approach enables the model to capture both the global scene structure and fine-grained local details, resulting in more accurate placement and scaling of 3D Gaussians when input views are limited.

Central to HDGS is a new loss function called \textit{Cascade Pearson Correlation Loss (CPCL)}. This loss compares depth patches at various scales using Pearson correlation, a scale-invariant measure of structural similarity. The loss is applied in a cascading fashion, where depth alignment begins at coarse resolution and progressively moves to finer levels. This structure-aware regularization enables more robust geometric learning, especially in the presence of depth noise, scale ambiguity, or missing regions in depth predictions.

We evaluate our method on the LLFF and DTU benchmarks and show that HDGS achieves state-of-the-art performance in sparse-view novel view synthesis. Our approach improves both structural fidelity and visual quality while maintaining efficient real-time rendering.

\noindent \textbf{Our main contributions are as follows:}
\begin{itemize}
\item \textbf{Hierarchical Depth-Guided Splatting (HDGS)}: We introduce a coarse-to-fine, multi-scale depth supervision framework for 3D Gaussian Splatting that progressively refines geometry under sparse views, improving Gaussian placement and structural fidelity.
\item \textbf{Cascade Pearson Correlation Loss (CPCL)}: We propose a correlation-based objective that enforces structural agreement between rendered depth and predicted depth priors across multiple patch scales, providing robust supervision under depth noise and scale ambiguity.
\item \textbf{State-of-the-art Performance}: We demonstrate consistent improvements on LLFF and DTU under sparse-view settings, achieving state-of-the-art performance compared with existing NeRF-based and 3DGS-based baselines.
\end{itemize}

\begin{figure*}[t]
    \centering
    \includegraphics[width=1\textwidth]{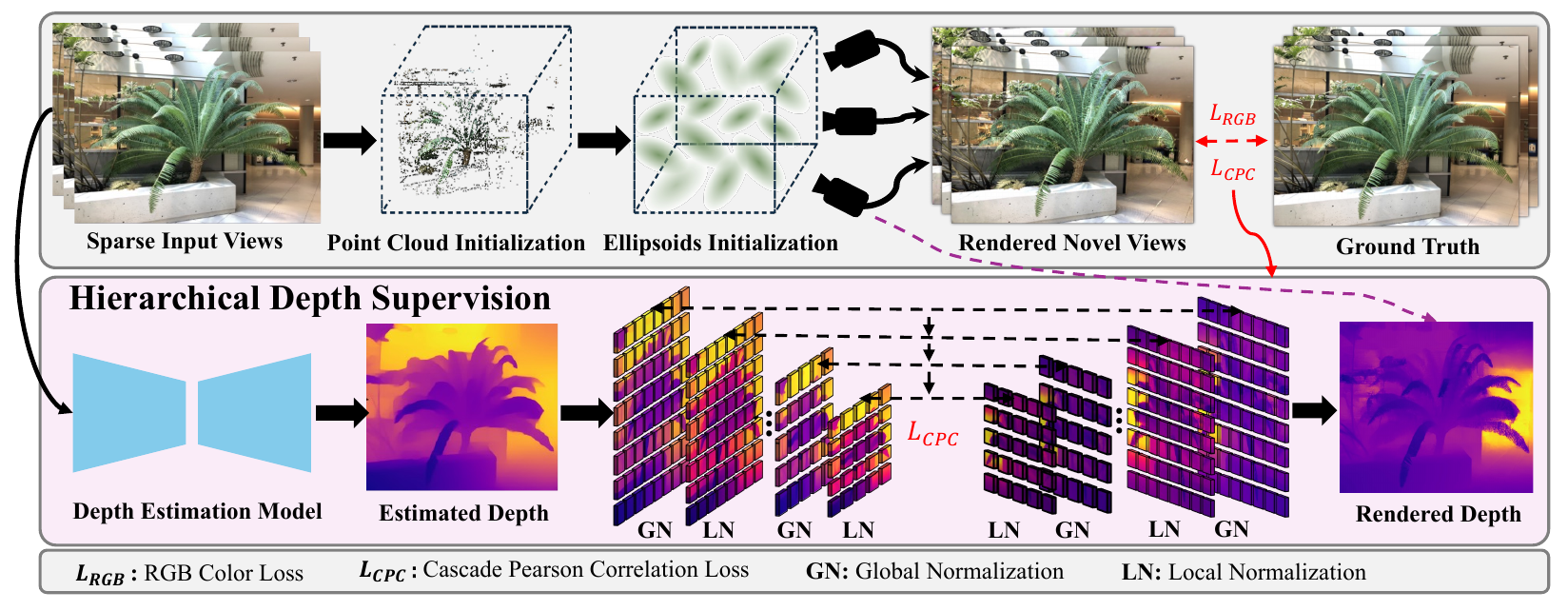}
    \caption{Overview of our framework: Given sparse input views, we initialize 3D Gaussians using a dense point cloud and supervise geometry via hierarchical alignment between rendered and predicted depth maps. A cascade Pearson correlation loss is applied across multi-scale patches and normalization modes, enabling accurate reconstruction under sparse-view conditions.}
    \label{fig:method}
\end{figure*}

\section{Related Work}
\label{sec:related_work}
\subsection{Radiance Fields for Novel View Synthesis}

Neural Radiance Fields (NeRF) have significantly advanced novel view synthesis by modeling scenes as implicit radiance fields with MLPs~\cite{mildenhall2021nerf}. Many extensions improve anti-aliasing, robustness, and efficiency~\cite{barron2021mip, barron2022mip, truong2023sparf, kim2023upnerf, fridovich2023kplanes}. However, NeRF-style methods still require dense volumetric sampling and expensive per-ray evaluations, which limit real-time deployment.

More recently, explicit representations have gained traction. Among them, 3D Gaussian Splatting (3DGS) models scenes with anisotropic 3D Gaussians and renders them via differentiable splatting~\cite{kerbl20233d}. This rasterization-friendly design provides high-quality reconstructions with real-time rendering, making 3DGS an attractive alternative to NeRF. However, 3DGS still struggles in sparse-view settings where geometric cues are limited, leaving substantial room for improvement in robustness and geometric fidelity under low view counts.

\subsection{Sparse View Novel View Synthesis}

Novel view synthesis from sparse inputs is challenging due to limited geometric supervision. NeRF-based methods address this with various regularization strategies, including geometric and appearance priors, frequency control, semantic consistency, and visibility-aware constraints~\cite{niemeyer2022regnerf, yang2023freenerf, jain2021putting, wang2023sparsenerf}. These approaches improve robustness but remain computationally demanding and are not tailored to the strengths of explicit 3DGS representations.

Within 3D Gaussian Splatting (3DGS) methods, several approaches leverage multi-view priors or monocular depth to guide optimization~\cite{han2024binocular, xu2024mvpgs, zhu2024fsgs, paliwal2024coherentgs, chung2024depth}, while others employ co-regularization strategies that train multiple Gaussian models to reduce overfitting and improve consistency~\cite{zhang2024cor, zhao2025self}. However,  depth-based approaches like ~\cite{li2024dngaussian} rely on global-scale depth predictions and often overlook fine local structures or suffer from scale inconsistencies. In contrast, our method introduces hierarchical depth supervision that integrates estimated depth cues at multiple granularities, enabling robust and consistent depth alignment across scales in sparse-view 3DGS.
\section{Methods}

\subsection{Preliminary of 3D Gaussian Splatting}

3D Gaussian Splatting (3DGS) explicitly represents a scene as a set of anisotropic Gaussian primitives. Each Gaussian is parameterized by a mean (center) position \( \boldsymbol{\mu} \in \mathbb{R}^3 \), a covariance matrix \( \boldsymbol{\Sigma} \in \mathbb{R}^{3 \times 3} \), an opacity \( \alpha \in \mathbb{R} \), and a color feature represented by spherical harmonics coefficients \( \mathbf{c} \in \mathbb{R}^{3(l+1)^2} \), where \( l \) is the degree of spherical harmonics.

The influence of each Gaussian at a position \( \mathbf{x} \in \mathbb{R}^3 \) in 3D space is modeled as:
\begin{equation}
G(\mathbf{x}) = \exp\left(-\frac{1}{2}(\mathbf{x} - \boldsymbol{\mu})^\top \boldsymbol{\Sigma}^{-1}(\mathbf{x} - \boldsymbol{\mu})\right).
\end{equation}

To ensure that \( \boldsymbol{\Sigma} \) remains positive semi-definite and  optimizable, it is decomposed into a rotation matrix \( \mathbf{R} \in \mathbb{R}^{3 \times 3} \) and a diagonal scaling matrix \( \mathbf{S} = \mathrm{diag}(s_1, s_2, s_3)  \) as:
\begin{equation}
\boldsymbol{\Sigma} = \mathbf{R} \mathbf{S}^2 \mathbf{R}^\top.
\end{equation}

During rendering, each Gaussian is projected from 3D world coordinates to 2D image coordinates using the camera's extrinsic matrix \( \mathbf{T}_{wc} = [\mathbf{R}_{wc} | \mathbf{t}_{wc}] \in \text{SE}(3) \) and intrinsic matrix \( \mathbf{K} \in \mathbb{R}^{3 \times 3} \). The mean and covariance in image space are computed as:
\begin{equation}
\boldsymbol{\mu}' = \pi(\mathbf{K} (\mathbf{R}_{wc} \boldsymbol{\mu} + \mathbf{t}_{wc})), \quad
\boldsymbol{\Sigma}' = \mathbf{J} \mathbf{R}_{wc} \boldsymbol{\Sigma} \mathbf{R}_{wc}^\top \mathbf{J}^\top,
\end{equation}
where \( \pi(\cdot) \) denotes the perspective projection function, and \( \mathbf{J} \in \mathbb{R}^{2 \times 3} \) is the Jacobian matrix of the affine approximation of the projection.

\subsection{Hierarchical Depth Supervision}

To improve geometric reconstruction under sparse-view constraints, we propose a hierarchical depth-guided mechanism that enforces consistency between the rendered depth \( D_r \in \mathbb{R}^{1 \times H \times W} \) and the predicted depth prior \( D_p \in \mathbb{R}^{1 \times H \times W} \) across multiple spatial resolutions. This multi-scale supervision scheme progressively aligns coarse scene geometry and fine-grained local structures, making it more robust to view sparsity and scale ambiguity.

We define a set of patch sizes \( \mathcal{S} = \{s_1, s_2, \ldots, s_N\} \) that represent varying levels of spatial granularity. For each scale \( s \in \mathcal{S} \), we divide both depth maps into non-overlapping patches of size \( s \times s \) via an unfolding operation. This process flattens each depth map \( D \in \mathbb{R}^{1 \times H \times W} \) into patch sets:
\begin{equation}
\mathcal{P}_r^s = \text{Unfold}(D_r, s), \:
\mathcal{P}_p^s = \text{Unfold}(D_p, s),
\end{equation}
\begin{equation}
\mathcal{P}_r^s = \left\{ \mathbf{x}_1^r, \ldots, \mathbf{x}_K^r \right\}, \:
\mathcal{P}_p^s = \left\{ \mathbf{x}_1^m, \ldots, \mathbf{x}_K^m \right\},
\end{equation}
where each \( \mathbf{x}_k^r, \mathbf{x}_k^m \in \mathbb{R}^{s^2} \) represents the vectorized pixel values of the \( k \)-th patch and \( K = \frac{H \cdot W}{s^2} \) is the number of patches per image at that scale.

\subsubsection{Local Normalization}

Local normalization focuses on internal patch structure while ignoring absolute depth scale. Each patch is normalized independently by subtracting its mean and dividing by its standard deviation:
\begin{equation}
\hat{\mathbf{x}}_k^r = \frac{\mathbf{x}_k^r - \mu(\mathbf{x}_k^r)}{\sigma(\mathbf{x}_k^r) + \epsilon}, \quad
\hat{\mathbf{x}}_k^m = \frac{\mathbf{x}_k^m - \mu(\mathbf{x}_k^m)}{\sigma(\mathbf{x}_k^m) + \epsilon}.
\end{equation}

This normalization removes local scale and bias effects, allowing the loss to focus purely on patch-wise structural similarity. The patch-level local depth loss is then computed as a normalized mean squared error between patches:
\begin{equation}
\mathcal{L}_{\text{local}}^s = \frac{1}{K} \sum_{k=1}^{K} \left\| \hat{\mathbf{x}}_k^r - \hat{\mathbf{x}}_k^m \right\|_2^2.
\end{equation}

\subsubsection{Global Normalization}

To complement the localized focus, we also incorporate global normalization, which enforces consistency in relative depth scale across the entire image. Instead of using per-patch statistics, we normalize each patch using a shared global standard deviation computed over the entire depth map:
\begin{equation}
\sigma_{\text{global}}^r = \text{std}(D_{\text{rendered}}), \quad
\sigma_{\text{global}}^m = \text{std}(D_{\text{mono}}),
\end{equation}
\begin{equation}
\tilde{\mathbf{x}}_k^r = \frac{\mathbf{x}_k^r - \mu(\mathbf{x}_k^r)}{\sigma_{\text{global}}^r + \epsilon}, \quad
\tilde{\mathbf{x}}_k^m = \frac{\mathbf{x}_k^m - \mu(\mathbf{x}_k^m)}{\sigma_{\text{global}}^m + \epsilon}.
\end{equation}

This formulation preserves scene-level depth relationships and mitigates inconsistencies in global geometry. The global depth loss is similarly computed as:
\begin{equation}
\mathcal{L}_{\text{global}}^s = \frac{1}{K} \sum_{k=1}^{K} \left\| \tilde{\mathbf{x}}_k^r - \tilde{\mathbf{x}}_k^m \right\|_2^2.
\end{equation}

\subsubsection{Multi-Scale Aggregation}

To fully exploit depth cues at multiple spatial frequencies, we combine both local and global losses at each scale \( s \) using a weighted sum:
\begin{equation}
\mathcal{L}_{\text{depth}}^s = w_{\text{local}} \cdot \mathcal{L}_{\text{local}}^s + w_{\text{global}} \cdot \mathcal{L}_{\text{global}}^s.
\end{equation}

Finally, we average the aggregated losses across all scales to obtain the total hierarchical depth supervision objective:
\begin{equation}
\mathcal{L}_{\text{depth}} = \frac{1}{|\mathcal{S}|} \sum_{s \in \mathcal{S}} \mathcal{L}_{\text{depth}}^s.
\end{equation}

This design provides supervision that is both spatially adaptive and scale-aware. Local normalization guides the model to capture fine-grained structures, while global normalization enforces coherent geometry over larger regions. Together, they refine both global layout and local details, improving reconstruction fidelity in sparse-view conditions.

% Monocular depth priors are typically uncalibrated, whereas depths rendered from 3DGS are in metric scale. 

\subsection{Cascade Pearson Correlation Loss}

Even with hierarchical depth supervision, directly aligning predicted and rendered depths is difficult due to scale mismatch. Monocular depth priors are typically uncalibrated,
whereas depths rendered from 3DGS are in metric scale. In this setting, standard \(\ell_1\) or \(\ell_2\) losses are unreliable because they penalize absolute differences rather than structural similarity.

We therefore adopt the Pearson correlation coefficient as a scale-invariant depth alignment loss. Given a rendered depth patch \( D_r = \{d_r^i\}_{i=1}^N \) and a corresponding prior patch \( D_p = \{d_p^i\}_{i=1}^N \), the Pearson correlation loss is:
\begin{equation}
\mathcal{L}_{\text{PCC}} = 1 - \frac{\sum_{i=1}^N (d_r^i - \bar{d}_r)(d_p^i - \bar{d}_p)}{\sqrt{\sum_{i=1}^N (d_r^i - \bar{d}_r)^2} \cdot \sqrt{\sum_{i=1}^N (d_p^i - \bar{d}_p)^2}},
\end{equation}
where \(\bar{d}_r = \frac{1}{N} \sum_{i=1}^N d_r^i\) and \(\bar{d}_p = \frac{1}{N} \sum_{i=1}^N d_p^i\). This loss focuses on relative depth ordering and structural similarity while being invariant to scale and offset.

To provide supervision at different spatial frequencies, we apply this loss over a set of patch sizes \(\mathcal{S} = \{s_1, s_2, \ldots, s_N\}\). For each patch size \( s \in \mathcal{S} \), the rendered and monocular depth maps are divided into non-overlapping patches. At each scale, we compute a combined patch loss:
\begin{equation}
\mathcal{L}_{\text{patch}}^{s} = \alpha \cdot \mathcal{L}_{\text{PCC}}^{s} + \beta \cdot \mathcal{L}_{\text{MSE}}^{s},
\end{equation}
where \(\mathcal{L}_{\text{PCC}}^{s}\) and \(\mathcal{L}_{\text{MSE}}^{s}\) denote the Pearson correlation and normalized mean squared error over patches of size \(s\), and \(\alpha, \beta\) balance the two terms.

The final Cascade Pearson Correlation Loss aggregates supervision over all patch sizes:
\begin{equation}
\mathcal{L}_{\text{CPCL}} = \frac{1}{|\mathcal{S}|} \sum_{s \in \mathcal{S}} \mathcal{L}_{\text{patch}}^{s},
\end{equation}
providing scale-invariant, multi-scale structural guidance that complements the hierarchical depth supervision.

\begin{table*}[!ht]
\caption{Experiment results on DTU~\cite{jensen2014large} and LLFF~\cite{mildenhall2019local} with 3 input views. 
The \colorbox[HTML]{9EBCAF}{first}, \colorbox[HTML]{DFE9E4}{second}, and \colorbox[HTML]{F2EDED}{third} best results are highlighted.}
\centering
\resizebox{\textwidth}{!}{
\begin{tabular}{lccccccc}
\toprule
\multirow{2}{*}{\textbf{Method}} & \multirow{2}{*}{\textbf{Setting}} & \multicolumn{3}{c}{\textbf{DTU}} & \multicolumn{3}{c}{\textbf{LLFF}} \\
\cmidrule(lr){3-5} \cmidrule(lr){6-8}
& & \textbf{PSNR} $\uparrow$ & \textbf{SSIM} $\uparrow$ & \textbf{LPIPS} $\downarrow$ & \textbf{PSNR} $\uparrow$ & \textbf{SSIM} $\uparrow$ & \textbf{LPIPS} $\downarrow$ \\
\midrule

PixelNeRF  &\multirow{7}{*}{NeRF-based Methods} & 16.82 & 0.695 & 0.270 & 7.93 & 0.272 & 0.682 \\
MVSNeRF\cite{chen2021mvsnerf}  & & 18.63 & 0.769 & 0.197 & 17.25 & 0.557 & 0.356 \\
Mip-NeRF\cite{barron2021mip}  & & 10.21 & 0.597 & 0.348 & 16.17 & 0.431 & 0.495 \\
DietNeRF\cite{jain2021putting}  & & 11.85 & 0.633 & 0.314 & 14.94 & 0.370 & 0.496 \\
RegNeRF\cite{niemeyer2022regnerf}   & & 18.89 & 0.745 & 0.190 & 19.08 & 0.587 & 0.336 \\
FreeNeRF\cite{yang2023freenerf}  & & 19.92 & 0.787 & 0.182 & 19.63 & 0.612 & 0.308 \\
SparseNeRF\cite{wang2023sparsenerf}  & & 19.55 & 0.769 & 0.201 & 19.86 & 0.624 & 0.328 \\

\midrule
3DGS\cite{kerbl20233d}  & \multirow{8}{*}{3DGS-based Methods} & 13.38 & 0.614 & 0.356 & 15.52 & 0.482 & 0.370 \\

FSGS\cite{zhu2024fsgs}  & & - & - & - & 20.43 & 0.682 & 0.248 \\

DNGaussian\cite{li2024dngaussian}  & & 18.23 & 0.780 & 0.184 & 18.86 & 0.600 & 0.294 \\

CoR-GS\cite{zhang2024cor}  & & 20.36 & 0.854 & 0.140 & 20.45 & \cellcolor[HTML]{F2EDED}0.712 & \cellcolor[HTML]{DFE9E4}0.196 \\

FewViewGS\cite{yin2024fewviewgs}  & & 19.74 & 0.861 & 0.127 & \cellcolor[HTML]{F2EDED} 20.54 & 0.693 & 0.214 \\

MVPGS\cite{xu2024mvpgs}  & & \cellcolor[HTML]{F2EDED}20.50 & \cellcolor[HTML]{DFE9E4}0.871 & \cellcolor[HTML]{9EBCAF}0.106 
& 20.39 & \cellcolor[HTML]{DFE9E4}0.715 & \cellcolor[HTML]{F2EDED}0.203 \\

SCGaussian\cite{peng2024structure}  & & \cellcolor[HTML]{DFE9E4}20.63 & \cellcolor[HTML]{F2EDED}0.868 & \cellcolor[HTML]{F2EDED}0.114 & \cellcolor[HTML]{DFE9E4}20.77 & 0.705 & 0.218 \\

\midrule
\textbf{HDGS (Ours)} & 
& \cellcolor[HTML]{9EBCAF}21.47 
& \cellcolor[HTML]{9EBCAF}0.875 
& \cellcolor[HTML]{DFE9E4}0.109 
& \cellcolor[HTML]{9EBCAF}21.31 
& \cellcolor[HTML]{9EBCAF}0.738 
& \cellcolor[HTML]{9EBCAF}0.178 \\
\bottomrule
\end{tabular}
}
\label{tab:dtu-llff-final}

\end{table*}

\section{Experiments}

\subsection{Datasets}
We conduct experiments on two real-world datasets, LLFF~\cite{mildenhall2019local} and DTU~\cite{jensen2014large}, which are standard benchmarks for novel view synthesis. LLFF contains forward-facing indoor and outdoor scenes with diverse geometry and appearance. Following prior work~\cite{li2024dngaussian,han2024binocular}, we use every eighth image as a test view, uniformly sample three training views from the remaining images to simulate sparse-view conditions, and downsample all images by a factor of 8. DTU consists of 124 indoor object-centric scenes. In line with previous methods~\cite{li2024dngaussian,han2024binocular}, we train with three images per scene, downsample images by a factor of 4, and apply the provided masks during evaluation to focus on foreground reconstruction quality.

\subsection{Evaluation Metrics}
For quantitative evaluation, we adopt three common metrics: Peak Signal-to-Noise Ratio (PSNR), Structural Similarity Index (SSIM), and Learned Perceptual Image Patch Similarity (LPIPS). PSNR and SSIM are metrics where higher values indicate better results, measuring pixel-level accuracy and structural resemblance, respectively. In contrast, LPIPS is a perceptual metric, with lower values indicating better perceptual similarity.

\subsection{Implementation Details}

Our method is implemented in PyTorch with CUDA acceleration on an NVIDIA A6000 GPU, and is built upon the open-source gsplat v1.3 framework~\cite{ye2025gsplat}. We generate dense 3D point clouds using a pre‐trained VGGSfM model~\cite{wang2024vggsfm}. This dense initialization provides strong geometric priors under sparse‐view conditions. In addition, we incorporate monocular depth priors using DepthAnything V2~\cite{yang2024depthanythingv2} to further guide depth supervision. All scenes in the LLFF and DTU datasets are trained for 30,000 iterations. Our loss function includes hierarchical depth supervision with both local and global consistency terms. The local depth loss is defined as $L_{\mathrm{local}} = w_{p}L_{\mathrm{PC}} + w_{l2}L_{2}$, with $w_{p}=0.1$ and $w_{l2}=0.9$, and the global depth loss is defined analogously. 
% These components are then combined with $w_{\mathrm{local}}=0.7$ and $w_{\mathrm{global}}=0.3$, and the hierarchical depth term is integrated into the final training objective with weight $\lambda=5\times10^{-3}$.  

\begin{figure*}[!t]
    \centering
    \includegraphics[width=\textwidth]{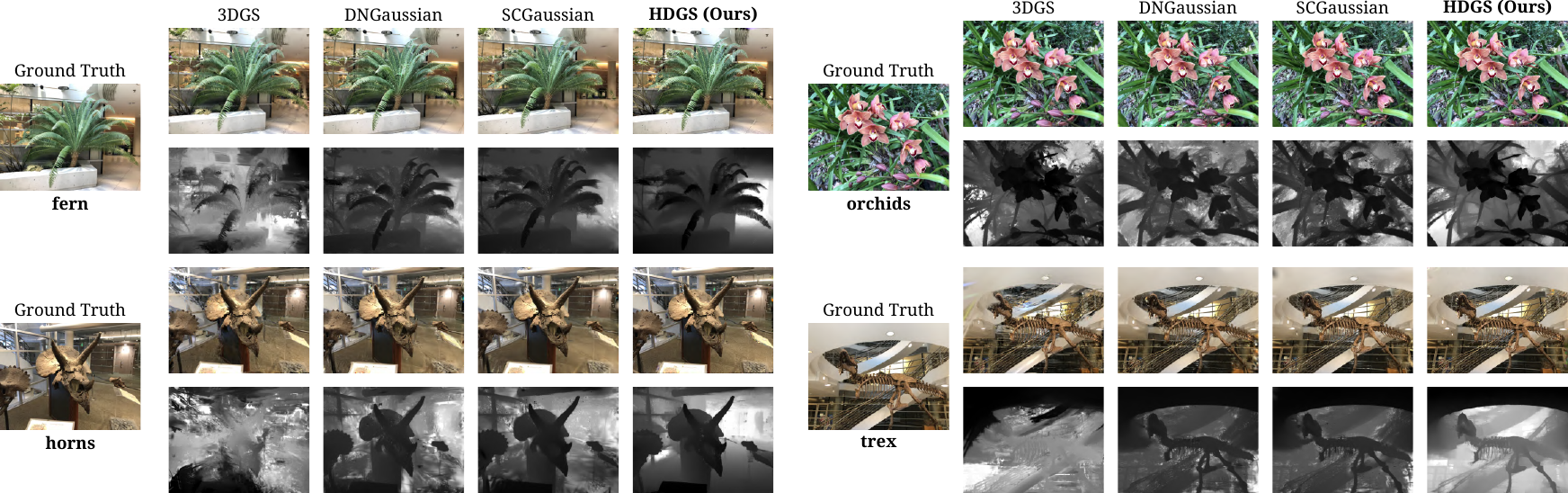}
    \caption{Qualitative comparison of rendered RGB images and depth maps on the LLFF dataset using 3 input views.}
    \label{fig:quan_LLFF}
\end{figure*}

\begin{figure*}[!t]
    \centering
    \includegraphics[width=\textwidth]{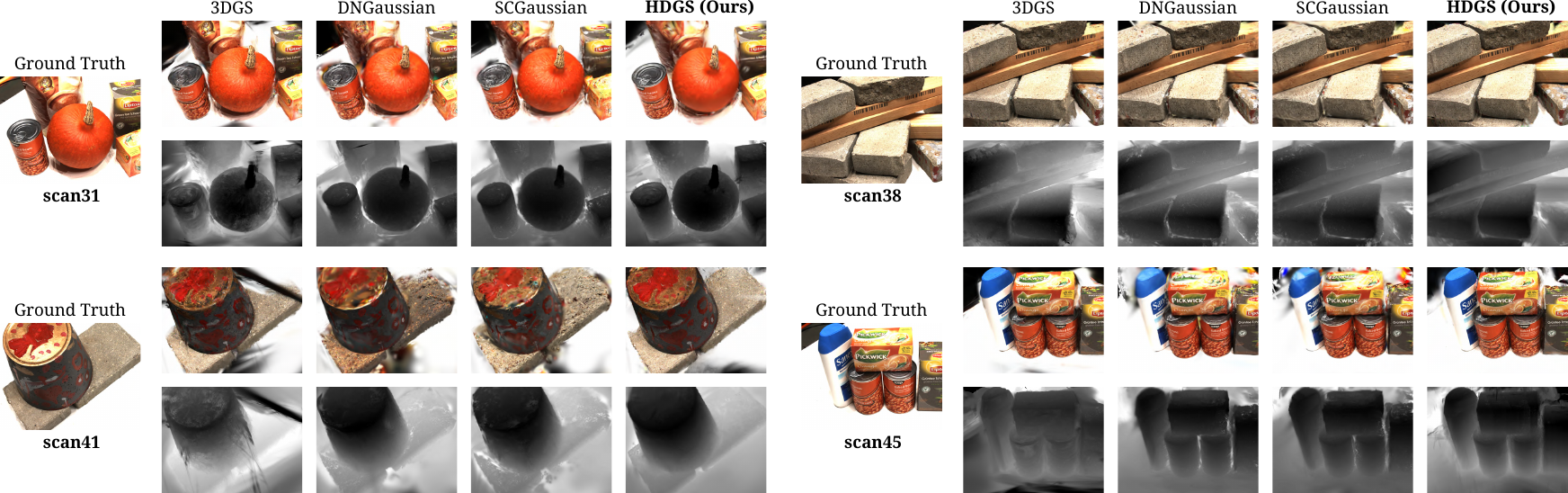}
    \caption{Qualitative comparison of rendered RGB images and depth maps on the DTU dataset using 3 input views.}
    \label{fig:quan_DTU}
\end{figure*}

\subsection{Baselines}
We compare HDGS against a comprehensive set of state-of-the-art sparse-view NVS methods. Our evaluation includes representative NeRF-based approaches~\cite{jain2021putting, niemeyer2022regnerf, yang2023freenerf, wang2023sparsenerf} and recent 3DGS-based techniques~\cite{xiong2023sparsegs, li2024dngaussian, zhu2024fsgs, xu2024mvpgs, peng2024structure, yin2024fewviewgs, kerbl20233d}. Together, these baselines cover both implicit volumetric and explicit point-based representations under sparse-view settings.

\subsection{Comparisons}

\subsubsection{LLFF}
% Table~\ref{tab:dtu-llff-final} shows that HDGS achieves the highest performance across all metrics on LLFF, outperforming both NeRF-based and 3DGS-based baselines. Compared to NeRF-based approaches, HDGS achieves sharper and more consistent renderings, overcoming the artifacts and inconsistencies that arise from implicit radiance field models under sparse supervision. Among 3DGS-based methods, HDGS also surpasses recent approaches, demonstrating that hierarchical depth supervision combined with cascade correlation loss provides stronger geometric constraints. 
Table~\ref{tab:dtu-llff-final} shows that HDGS achieves the best performance on LLFF across all metrics, outperforming both NeRF-based and 3DGS-based methods. It produces sharper and more consistent renderings under sparse supervision, demonstrating the effectiveness of hierarchical depth supervision and cascade correlation loss.
Figure~\ref{fig:quan_LLFF} presents qualitative comparisons on LLFF scenes. In \textit{fern}, baseline methods produce blurry artifacts and color bleeding, while HDGS generates crisp leaf contours and improved depth layering. In \textit{horns}, baseline outputs artifacts on the floor and mirror reflection area, while HDGS retains their curved geometry with minimal distortion. For \textit{trex}, HDGS eliminates ghosting and blurred parts of the scene, producing sharper silhouettes and textures with improved background separation.

\subsubsection{DTU}
As shown in Table~\ref{tab:dtu-llff-final}, HDGS also achieves the best overall performance on DTU, outperforming both NeRF-based and 3DGS-based baselines in terms of fidelity and perceptual quality. Figure~\ref{fig:quan_DTU} shows qualitative comparisons on representative DTU scenes. In \textit{scan31}, which contains thin metal structures and complex occlusions, baselines produce soft and distorted edges, while HDGS reconstructs fine rods and layered geometry with high fidelity. In \textit{scan41}, where depth discontinuities are prominent, HDGS reduces blending artifacts and recovers stronger depth contrast. Finally, in \textit{scan45}, HDGS produces sharper edges and cleaner geometry, avoiding the color spill and warping seen in other methods. 

\subsection{Ablation Study}

\subsubsection{Effect of Key Components}

\begin{table}[!t]
\caption{Ablation study of individual components, including Hierarchical Depth (HD) supervision, Pearson Correlation (PC) loss, and Dense Initialization (DI), on the LLFF dataset.}
\centering
\begin{tabular}{lccc}
\toprule
\textbf{Method} & \textbf{PSNR} $\uparrow$ & \textbf{SSIM} $\uparrow$ & \textbf{LPIPS} $\downarrow$ \\
\midrule
3DGS & 15.52 & 0.482 & 0.370 \\
\midrule
DI & 19.98 & 0.667 & 0.238 \\
DI + PC & 20.32 & 0.707 & 0.226 \\
DI + HD & 20.84 & 0.729 & 0.213 \\
DI + HD + PC & \textbf{21.31} & \textbf{0.738} & \textbf{0.178} \\
\bottomrule
\end{tabular}
\label{tab:ablation_components}
\vspace{-4mm}
\end{table}

Table~\ref{tab:ablation_components} reports an incremental ablation on LLFF under sparse-view settings. Switching from the default sparse SfM point cloud to Dense Initialization (DI) substantially improves all metrics, indicating that sparse SfM is often too incomplete and noisy to provide reliable Gaussians with few views. Adding PC on top of DI further improves performance, with a notable gain in PSNR and SSIM, suggesting that correlation-based alignment provides additional structural guidance beyond initialization quality. Adding HD supervision to DI yields larger improvements than PC alone, consistent with multi-scale depth alignment better constraining both global geometry and local details. Finally, combining HD and PC achieves the best results, demonstrating that PC complements HD by improving perceptual consistency once geometry is better constrained.

\subsubsection{Effect of Patch Configurations}
We evaluate the impact of different patch scale configurations in the HD+PC module (Table~\ref{tab:ablation_patch_sizes}). Moving from single-scale to multi-scale supervision  improves performance, and the best results are achieved with three scales (4+8+16). Small patches emphasize fine surface details, while mid-scale patches provide sufficient context for stable depth alignment. Importantly, extending the hierarchy with an additional coarse scale slightly degrades performance. We attribute this to the diminished discriminative power of very large patches, which average out fine geometric variation and introduce redundancy. Based on this trade-off, we adopt the three-scale hierarchy in all experiments.

\begin{table}[!t]
\caption{Effect of patch-scale configurations in the HD+PC module on LLFF (3 views).}
\label{tab:ablation_patch_sizes}
\centering
\small
\setlength{\tabcolsep}{3pt}
\renewcommand{\arraystretch}{1.05}

\begin{tabular}{l c r r r}
\toprule
\textbf{Patch scales} & \textbf{\#} & \textbf{ PSNR} $\uparrow$ & \textbf{ SSIM} $\uparrow$ & \textbf{ LPIPS} $\downarrow$ \\
\midrule
4               & 1 &20.32 &0.678 &0.236 \\
4+8             & 2 & 20.71 & 0.707 & 0.212 \\
4+8+16          & 3 & \textbf{21.31} & \textbf{0.738} & \textbf{0.178} \\
4+8+16+32       & 4 & 21.11 & 0.720 & 0.186 \\
\bottomrule
\end{tabular}
\vspace{-0.5em}
\end{table}

\section{Conclusion}
In this paper, we present Hierarchical Depth-Guided Sparse-View 3D Gaussian Splatting for improved novel view synthesis under sparse-view conditions. Our hierarchical depth supervision refines geometry from coarse to fine scales, and the Cascade Pearson Correlation Loss robustly aligns rendered and estimated depths. Experiments on standard benchmarks show that HDGS achieves state-of-the-art geometric fidelity, rendering quality, and robustness in sparse-view scenarios.

\section{Acknowledgment}
This work was partially supported by the ECU Early-Mid Career Researcher (EMCR) Grant Scheme.

\bstctlcite{BSTcontrol}
\bibliographystyle{IEEEtran}


% \bibliography{references}

\end{document}